\documentclass[letterpaper, 10 pt, conference]{ieeeconf}  

\IEEEoverridecommandlockouts                              

\usepackage{cite}
\usepackage{amsmath,amssymb,amsfonts}
\usepackage{algorithmic}
\usepackage{graphicx}
\usepackage{textcomp}
\usepackage{xcolor}

\usepackage{makeidx}

\usepackage{pgfplots}
\pgfplotsset{compat=newest}
\usepackage{multirow}
\usepackage{tabularx}
\usepackage{threeparttable}
\usepackage{booktabs}
\usepackage{array}
\usepackage{float}
\usepackage{graphbox}
\usepackage{subcaption}
\usepackage{tikz,tikz-3dplot}
\usetikzlibrary{fit,arrows,arrows.meta,automata,backgrounds,calc,chains,%
decorations.markings,decorations.pathreplacing,decorations.pathmorphing,%
matrix,positioning,shapes,shapes.geometric,shapes.symbols,spy,trees,tikzmark}
\usepackage{balance}
\usepackage[binary-units=true,product-units=single,per-mode=symbol,range-units=single,range-phrase=\,--\,,detect-all]{siunitx}
\DeclareSIUnit\pixel{px}
\usepackage{bm}
\usepackage{acronym}
\usepackage{tablefootnote}
\usepackage{hyperref}

\usepackage{caption}
\let\labelindent\relax
\usepackage[inline]{enumitem}

\let\vec\bm

\newcommand{\etal}{et al.~}

\usepackage{adjustbox}
\newcolumntype{R}[2]{%
    >{\adjustbox{angle=#1,lap=\width-(#2)}\bgroup}%
    l%
    <{\egroup}%
}
\newcolumntype{L}[1]{>{\raggedright\let\newline\\\arraybackslash\hspace{0pt}}m{#1}}

\title{\LARGE \bf
SLCF-Net: Sequential LiDAR-Camera Fusion for Semantic Scene Completion using a 3D Recurrent U-Net
}

\author{Helin Cao and Sven Behnke
\thanks{
	This research has been supported by MBZIRC price money.	All authors are with the Autonomous Intelligent Systems group, 
	Computer Science Institute VI -- Intelligent Systems and Robotics -- and the Center for Robotics and the Lamarr Institute for Machine Learning and Artificial Intelligence, University of Bonn, Germany;
        {\tt\small caoh@ais.uni-bonn.de}}%
}

\begin{document}

\maketitle
\thispagestyle{empty}
\pagestyle{empty}

\begin{abstract}
We introduce SLCF-Net, a novel approach for the Semantic Scene Completion (SSC) task that sequentially fuses LiDAR and camera data. It jointly estimates missing geometry and semantics in a scene from sequences of RGB images and sparse LiDAR measurements. The images are semantically segmented by a pre-trained 2D U-Net and a dense depth prior is estimated from a depth-conditioned pipeline fueled by Depth Anything. To associate the 2D image features with the 3D scene volume, we introduce Gaussian-decay Depth-prior Projection (GDP). This module projects the 2D features into the 3D volume along the line of sight with a Gaussian-decay function, centered around the depth prior. Volumetric semantics is computed by a 3D U-Net. We propagate the hidden 3D U-Net state using the sensor motion and design a novel loss to ensure temporal consistency. We evaluate our approach on the SemanticKITTI dataset and compare it with leading SSC approaches. The SLCF-Net excels in all SSC metrics and shows great temporal consistency.
\end{abstract}

\section{Introduction}
\label{sec:Introduction}
3D Semantic Scene Completion (SSC) aims to simultaneously estimate the complete geometry and semantics of a scene from sensor data, a task that has garnered increased attention in the computer vision and robotics community. Many existing methods still rely on single sensor input, such as RGB images or depth data (e.g., occupancy grids, point clouds, depth maps, etc.). Both RGB images and depth data can provide valuable information about the environment. However, intuitively, RGB images enable a dense interpretation of the semantic content with high spatial resolution, while depth data provides the scene geometry. The complementary nature of these two data modalities can facilitate SSC. Although RGB-D cameras directly provide both color and depth information, their short depth range (typically less than \SI{5}{\meter}) limits their application in outdoor urban scenarios. For autonomous vehicles, a common alternative is the combination of RGB images and LiDAR scans~\cite{geiger2012cvpr, sun2020scalability}. Addressing this prevalent configuration, we propose a novel framework that leverages both RGB and LiDAR data.

\begin{figure}[t]
    \centering
    \begin{subfigure}[b]{0.23\textwidth}
        \includegraphics[width=\textwidth]{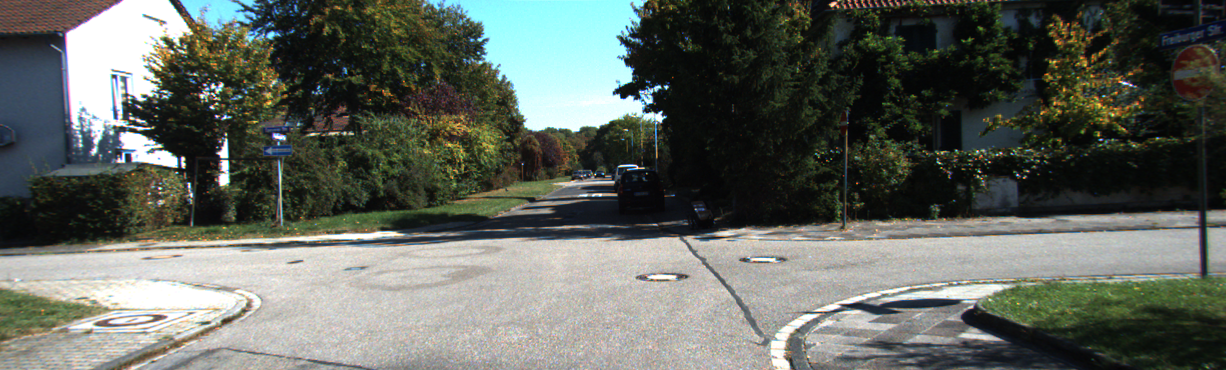}
        \caption{RGB input}
        \label{fig:rgb}
    \end{subfigure}
    \hfill
    \begin{subfigure}[b]{0.23\textwidth}
        \includegraphics[width=\textwidth]{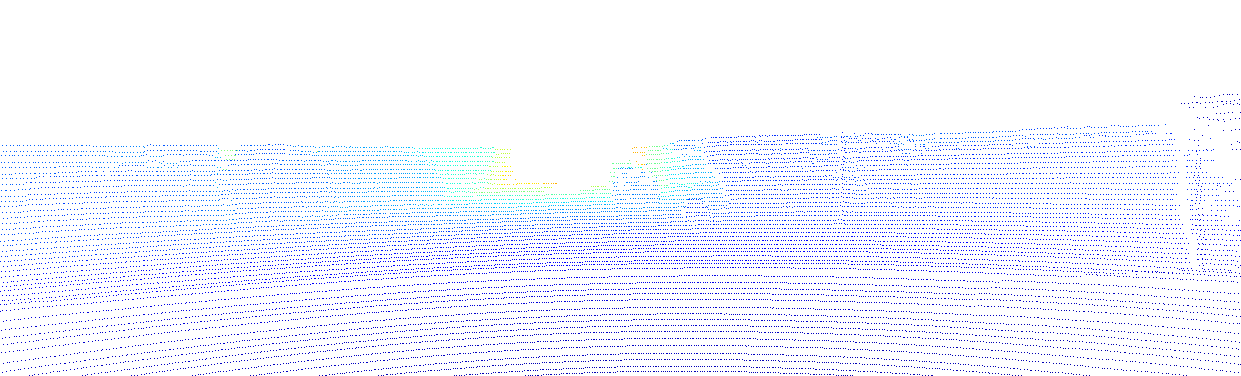}
        \caption{Sparse LiDAR depth input}
        \label{fig:depth}
    \end{subfigure}
    \vspace{1em}
    \begin{subfigure}[b]{0.23\textwidth}
        \includegraphics[angle=90, height=0.17\textheight, width=\textwidth]{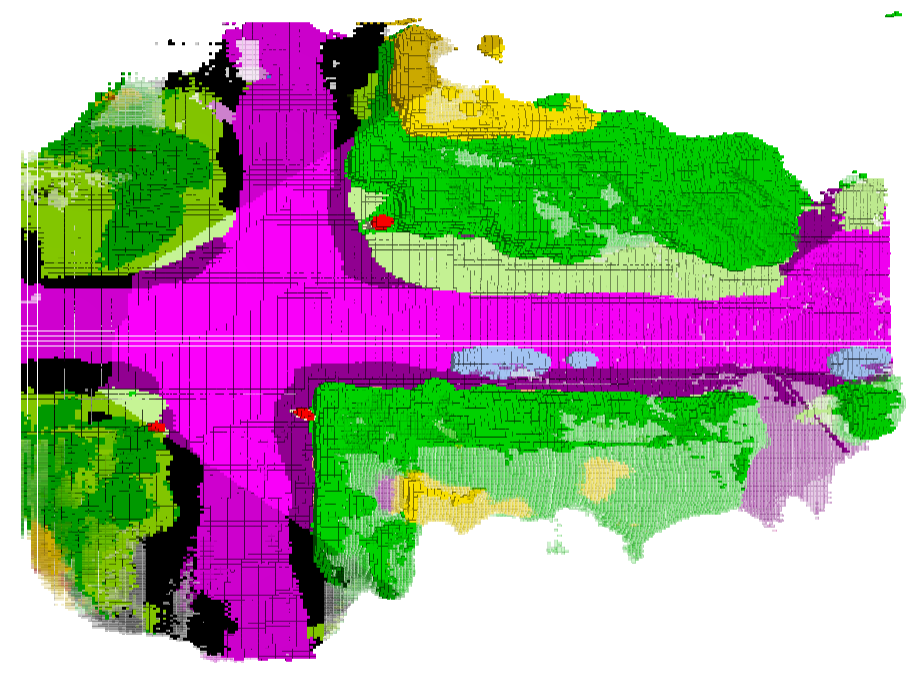}
        \caption{Dense semantic estimation}
        \label{fig:pred}
    \end{subfigure}
    \hfill
    \begin{subfigure}[b]{0.23\textwidth}
        \includegraphics[angle=90, height=0.17\textheight, width=\textwidth]{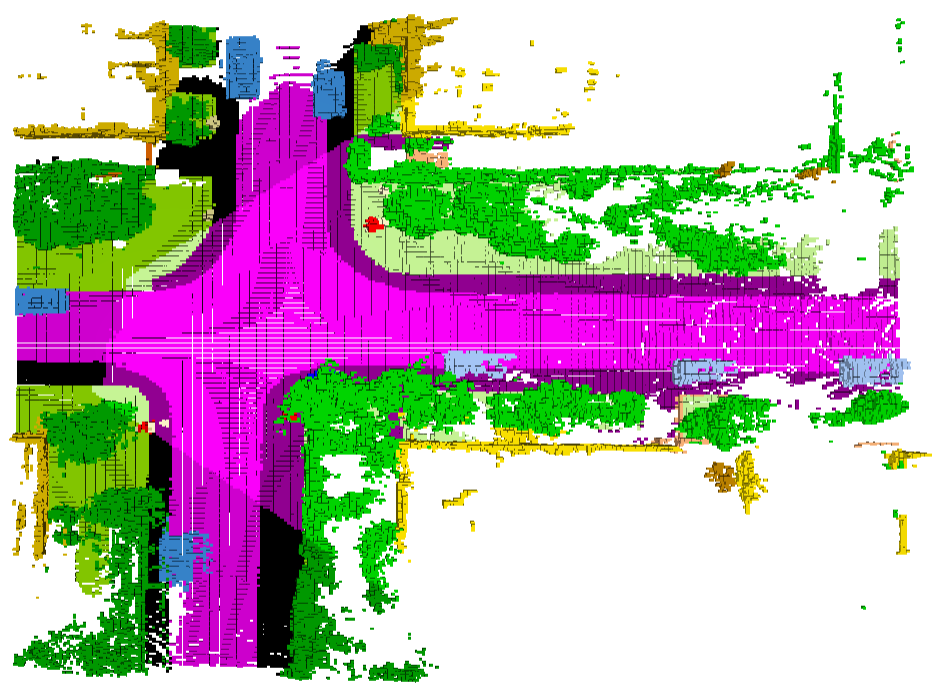}
        \caption{Ground-truth}
        \label{fig:gt}
    \end{subfigure}
    	\vspace{-.7em}
    \caption{SLCF-Net estimates the dense semantic scene as shown in (c) using sequences of RGB images (a) and aligned sparse LiDAR depth maps (b). Both (c) and (d) depict a voxelized scene as defined by the SemanticKITTI Benchmark~\cite{behley2019iccv} from the bird's-eye view. Parts of the scene in both the estimation (c) and the ground-truth (d) lie outside of the field of view (FoV), which are visualized as shadow areas. The unknown areas, as defined by the ground truth, are visualized at $20\%$ opacity in (c).}
    \label{fig:teaser}
    \vspace{-1.5em}
\end{figure}

In this work, we introduce SLCF-Net, designed to fuse sequences of RGB images and sparse 3D LiDAR scans to infer a 3D voxelized semantic scene. Fig.~\ref{fig:teaser} illustrates the inputs and outputs for urban driving scenarios. The RGB image, sourced from the KITTI dataset\cite{geiger2012cvpr}, is paired with a sparse depth map, projected from a calibrated LiDAR scan. The SLCF-Net then recursively estimates dense occupancy and semantic labels within the 3D scene volume.

Lifting a 3D scene from a 2D image is a well-known ill-posed problem~\cite{fahim2021single}. However, depth values provide crucial priors during scene reconstruction. To enhance the density of the depth map, we utilized the power of Depth Anything Model~\cite{yang2024depth}, which densely estimates relative distance from an RGB image. Then we scale the relative distance based on the raw sparse depth input to obtain the dense and absolute depth estimation. The dense depth estimation is utilized to project 2D features into the 3D volume. To model sensor noise and inference uncertainty, we smooth the projection with a depth-dependent function. Along the line of sight, as the distance from the depth prior increases, the weight associated with a voxel diminishes. This decrease is effectively captured using a Gaussian-decay function.

To the best of our knowledge, existing SSC methods reconstruct the scene from data in the current frame without considering historical information. Considering the scanning process in an autonomous vehicle, there is a large range of overlapping volumes between successive frames. Thus, the SSC model can benefit from leveraging previous frames. With the known poses of the frames, we propagate the 3D semantic scene representation from previous frames to compensate for the sensor motion. Benefiting from the ability of CNNs to preserve the spatial ordering of data, this alignment can be utilized not only in the output layer but also for feature fusion in latent space. We store the propagated previous feature as the hidden state and concatenate it with the current feature. We also designed an inter-frame consistency loss to enforce temporal consistency in the overlap area.

\noindent In summary, the main contributions of this paper include:
\begin{itemize}
  \item introduction of SLCF-Net: a novel SSC method that fuses 2.5D sparse depth maps with 2D RGB images, adaptable to various sensor configurations;
  \item the GDP module: a Gaussian-decay Depth-prior Projection method for projecting 2D features into 3D;
  \item a mechanism that propagates features from the previous frame to the current frame based on the known coordinate transformation.
\end{itemize}

\section{Related Work}
\label{sec:Related_Work}
\subsection{3D Semantic Scene Completion (SSC)}
Research on inferring missing geometry based on existing information has spanned decades. Traditional approaches typically involved filling small holes by employing continuous energy minimization, such as Poisson surface reconstruction~\cite{kazhdan2006poisson, kazhdan2013screened}. However, the advent of deep neural networks marked a paradigm shift. SSCNet~\cite{song2017semantic}, for instance, became the first to jointly infer an entire scene's geometry and semantics from a monocular depth image, defining the 'SSC' task as requiring joint inference of geometry and semantics. Over the years, various data types have been employed as inputs for SSC methods, as meticulously documented in a survey by Roldao et al.~\cite{roldao20223d}. Most current works utilize inputs such as depth~\cite{li2019depth}, occupancy grids~\cite{wu2020scfusion, roldao2020lmscnet}, point clouds~\cite{yan2021sparse, cheng2021s3cnet, zhong2020semantic}, and RGB~\cite{cai2021semantic, cherabier2018learning, cao2022monoscene}. In this work, we address SSC by utilizing a unique combination of inputs: an RGB image coupled with a sparse depth map---an approach particularly tailored for the sensor configuration of autonomous vehicles.

\subsection{Sensor Fusion}
Sensor fusion is a technique that combines data from diverse sensory sources to enhance the perception and understanding of the environment. Numerous studies focus on the fusion of camera and LiDAR data to achieve refined 3D reconstructions or precise pose estimations. Riegler \etal \cite{riegler2017octnetfusion} proposed a method to combine the high-resolution color information from cameras with LiDAR depth data using a voxel-based format, improving 3D reconstruction accuracy. Czarnowski \etal \cite{czarnowski2020deepfactors} introduced a keyframe-based SLAM method that fuses camera and LiDAR for robust pose estimation in dynamic environments. Bultmann \etal addressed multiple tasks by the fusion of smart edge sensors with overlapping fields of view, like human pose tracking~\cite{bultmann20223d}, robot pose estimation~\cite{bultmann2023external}, and dynamic object tracking~\cite{hau2022object}. They also introduced a UAV system tailored for real-time semantic interpretation, integrating multiple sensor modalities such as LiDAR, RGB-D camera, and thermal camera~\cite{bultmann2023real}.

\subsection{Sequence Learning}
Sequence learning involves learning from previous elements to support the understanding of current elements or the prediction of future elements. Sequence learning can help the model produce results based on historical information. In SSC, this approach will help improve temporal consistency. Historically, this field was primarily centered around machine translation. Classical architectures like RNN~\cite{medsker2001recurrent}, GRU~\cite{cho2014learning}, and LSTM~\cite{hochreiter1997long} emerged during this early era. In recent years, sequence learning expanded to video understanding. Simonyan \etal \cite{simonyan2014two} presents a two-stream network that divides the video into spatial and temporal streams. While the spatial stream focuses on individual frame processing to capture appearance attributes, the temporal stream interprets optical flow fields between frames, targeting motion dynamics comprehension. Villar-Corrales \etal \cite{villar2022mspred} proposed Multi-Scale Hierarchical Prediction (MSPred), a video prediction framework capable of forecasting video on multiple levels of abstraction and spatio-temporal granularity. By decomposing scenes into objects and modeling their temporal dynamics and relations, sharp multi-step predictions can be learned~\cite{VillarCorralesWB:ICIP23}. Expanding upon the videos, scene flow delves deeper by modeling the 3D\,+\,temporal 4D space. As an example, BEVDet4D~\cite{huang2022bevdet4d} explicitly maps the result of the previous frame to the current frame through a feature alignment operation, assuming the poses are known. This alignment simplified learning by compensating for ego-motion. Inspired by this, to utilize the historical information, we propagate the features from the previous frame to the current frame via coordinate transformation. Therefore, the model uses the features of the previous frame as a prior to help estimate the current frame. Benefiting from CNN's property of maintaining spatial order, this method can be used not only for the output layer but also for the latent scene representations.

\section{Method}
\label{sec:Method}
\begin{figure*}[!ht]
	\centering
	\includegraphics[width=0.9\textwidth]{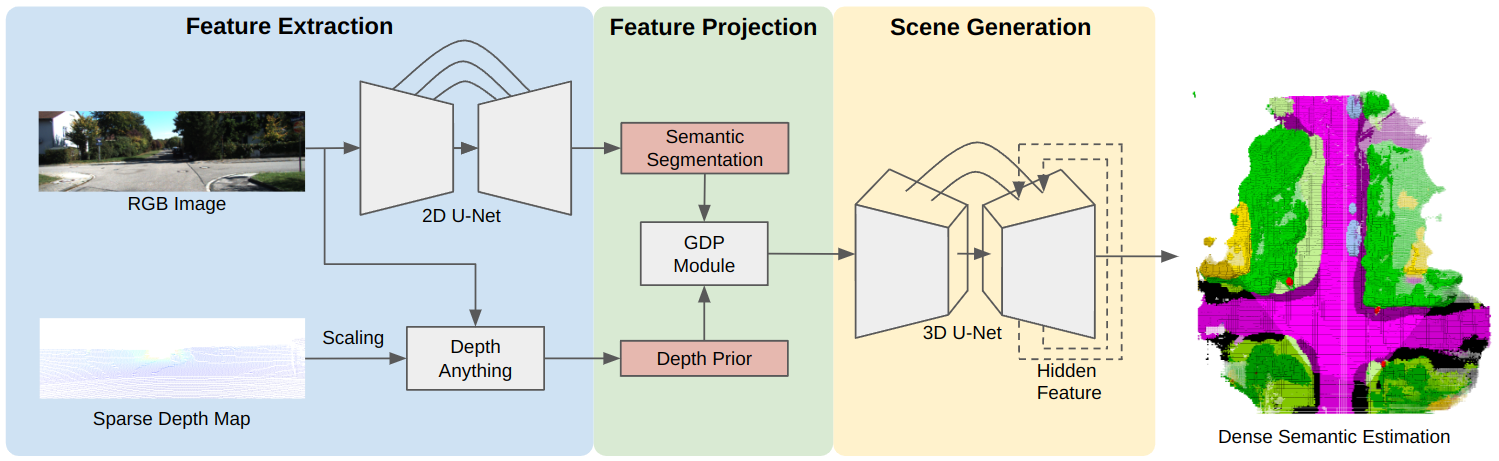}
	\caption{Overall pipeline of SLCF-Net. Given input sequences consisting of RGB images and a sparse depth map projected from a single sweep point cloud, the process is initiated by extracting the image feature into two channels. The 2D semantic features are extracted by an EfficientNet~\cite{tan2019efficientnet} with noisy student training~\cite{xie2020self}, while the relative depth is estimated by the Depth Anything Model. The relative depth is then scaled based on the sparse depth input to generate depth prior of entire image. Afterward, the Gaussian-decay Depth-prior Projection (GDP) module distributively back-projects the 2D features onto a predefined 3D volume using the depth priors. The 3D features are then fed into a 3D recurrent U-Net, which enables the harness of information from the previous frame. Finally, a dense grid semantic scene is generated as a comprehensive understanding of the environment.}
	\label{fig:SLCF-Net}
	\vspace{-1.5em}
\end{figure*}

3D Semantic Scene Completion aims to jointly estimate both the geometry and semantics of a 3D scene by assigning each voxel a label $L = {l_0, l_1,\ldots, l_M}$,  with M semantic classes and $l_0$
being empty space. To solve this task, SLCF-Net processes sequences of RGB images with associated sparse LiDAR depth, as illustrated in Fig.~\ref{fig:SLCF-Net}. Our method first computes 2D semantic features and dense depth priors from two input channels, respectively. The 2D semantic features are extracted by an EfficientNet~\cite{tan2019efficientnet} with noisy student training~\cite{xie2020self}. The relative depth is estimated by the Depth Anything Model and then scaled with raw depth input. Then the 2D semantic features are projected into a 3D voxel grid by the Gaussian-decay Depth-prior Projection (GDP) module. A 3D U-Net learns to generate the complete semantic scene from the 3D features, using a prior propagated from the semantic scene representation of the previous frame compensated for sensor motion.

\subsection{2D-3D Feature Projection}
\label{sec:gdp}
Neural networks have shown prowess in autonomously discerning feature correlations, as exemplified in numerous image-to-image studies. However, the incorporation of extra dimensions in 3D data increases computational requirements, posing a significant challenge to understanding the relationship between 2D and 3D features. Our approach is inspired by the Lift, Splat, Shoot (LSS) mechanism~\cite{philion2020lift}: given known calibration, 2D features can be backprojected into a 3D volume along the line of sight. This inductive bias facilitates learning. Considering a scanning sequence with $N$ frames, each local frame is anchored to the camera coordinate $C_i$, where $i = 1,\ldots,N$. The SemanticKITTI dataset provides the intrinsic calibration and pose estimation of each frame. i.e., we assume the transformation $\vec{T}^{C_i}_W \in \mathbb{R}^{4\times4}$ from world to camera coordinates and the projection $\Pi_i(\cdot)$ to the image plane of $C_i$ to be known. According to the perspective projection model, a 3D point $\vec{P} = [X, Y, Z]^\top \in \mathbb{R}^{3}$ can be projected to a pixel $\vec{p} = [x, y]^\top \in \mathbb{R}^{2}$ as follows:
\begin{align}
\tilde{\vec{p}} = \Pi_i \vec{T}^{C_i}_W \tilde{\vec{P}}.
\label{eqn:project}
\end{align}
Here, $\tilde{\vec{P}} = [X, Y, Z, 1]^T$ and $\tilde{\vec{p}} = [x, y, 1]^T$ represent the homogeneous coordinates of $\vec{P}$ and $\vec{p}$, respectively. With Eq.~\ref{eqn:project}, all the voxels in the field of view can be associated with the corresponding 2D feature.

To back-project a pixel to a 3D point, the depth $d$ is necessary. In the SLCF-Net, the depth prior $\hat{d}$ is derived from the relative depth estimation of Depth Anything and scaled according to the LiDAR measurement. Thus, the back-projection is as follows:
\begin{align}
\hat{\tilde{\vec{P}}} = \vec{T}^W_{C_i} \hat{d} \Pi_i^{-1}\tilde{\vec{p}}.
\label{eqn:backproject}
\end{align}

For any given pixel coordinate $\vec{p}$, the model will estimate a 3D point $\vec{P}$ based on the depth prior $\hat{d}$. It is worth noticing that by construction the real 3D point is on the line of sight, but the depth prior $\hat{d}$ is based on sensor noise and inference inaccuracies and is hence only a plausible estimate. To model the uncertainty of the depth prior, we employ the Gaussian-decay function:
\begin{align}
f(\vec{P}|\hat{\vec{P}},\sigma) = e^{-\frac{||\vec{P}-\hat{\vec{P}}||^2}{2\sigma^2}}.
\label{eqn:gaussian}
\end{align}

\begin{figure}
    \centering
    \begin{subfigure}[b]{0.28\textwidth}
        \includegraphics[width=1.0\textwidth]{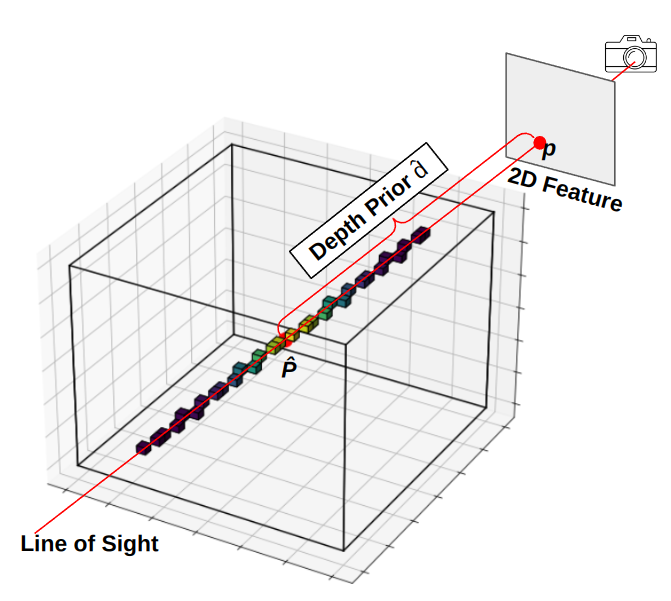}
        \caption{Backprojection with depth prior}
        \label{fig:gdpa}
    \end{subfigure}
    \hspace{0.05cm} 
    \begin{subfigure}[b]{0.19\textwidth}
        \includegraphics[width=\textwidth]{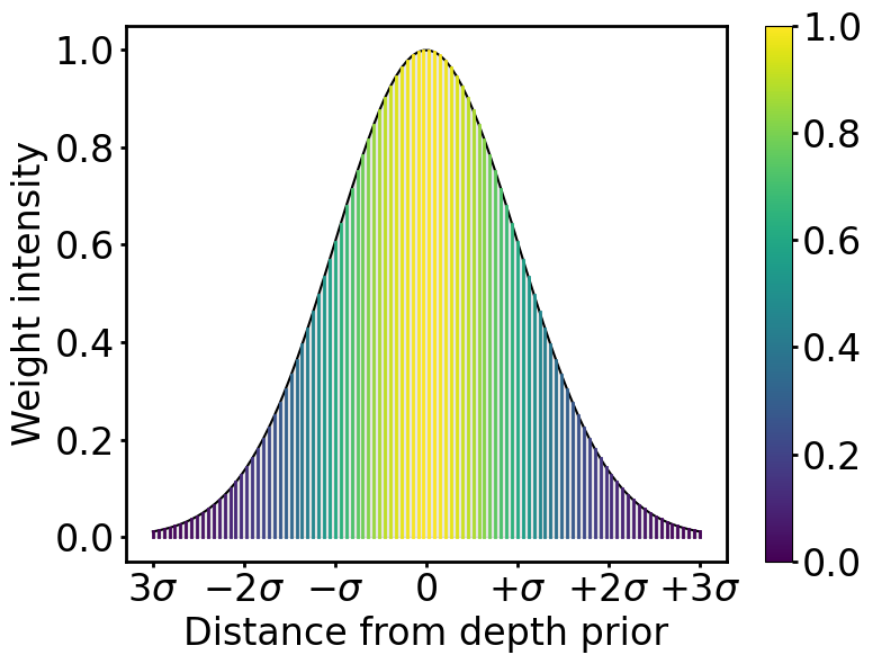}
        \caption{Gaussian-decay function}
        \label{fig:gdpb}
    \end{subfigure}
    \caption{Gaussian-decay Depth-prior Projection (GDP). (a) a 2D feature located at pixel coordinate $\vec{p}$ is projected to voxels in the 3D volume, following the line of sight. Using the depth prior $\hat{d}$, $\vec{P}$ is considered the most probable point and serves as the center of the Gaussian-decay function; (b) Gaussian-decay function weight.}
    \label{fig:gdp}
    \vspace{-1.em}
\end{figure}

Here, $\vec{P}$ is a voxel passed by the line of sight. $\hat{\vec{P}}$ is the estimated point from depth prior $\hat{d}$, which is also the center of the Gaussian-decay function. As illustrated in Fig.~\ref{fig:gdp}, the 2D feature is projected onto all voxels along the line of sight with Gaussian weight. This weight peaks at $\hat{\vec{P}}$ and decays with increasing distance from $\hat{\vec{P}}$. 

In the Gaussian-decay function, the parameter $\sigma$ signifies our confidence in the depth prior. When there is a significant discrepancy in the depth prior, a larger $\sigma$ enables the model to jump out from local optima. On the other hand, when the depth prior is accurate, a smaller $\sigma$ should be adopted to expedite the convergence of the 2D features to the real voxel. Since the depth information is derived from the depth-conditioned Depth Anything pipeline, the choice of $\sigma$ mainly depends on its noise distribution. The efficacy of the model can be enhanced with an optimal selection of $\sigma$, a factor that we will delve into in the ablation studies.

\subsection{Temporal Feature Propagation}
\label{sec:interframe}
A few classic frameworks, such as RNN, LSTM, and GRU, have been proposed for sequence learning and widely used in machine translation. RNNs were foundational for sequence learning, and the LSTM and GRU evolved to address gradient instability, letting networks manage longer sequences. To design a framework for temporal feature fusion, we analyzed the characteristics of the SSC task as follows. Although LSTM and GRU excel at extracting high-level features from long sequences, in SSC, the estimation is more dependent on current or nearby frames rather than a long history. Therefore, the primary advantage of LSTM is not needed for SSC tasks. Furthermore, considering that the goal of SSC is to estimate the occupancy and semantics of each voxel, the spatial structure and details are crucial. However, the gate mechanism in LSTM not only destroys the features' spatial structure but also drops details. The 3D U-Net, with its ability to maintain the spatial structure and details in the latent space, is well-suited for SSC. We propose a mapping for the 3D U-Net that propagates the detailed 3D semantic scene representation from the preview frame to the next frame while compensating for sensor motion. 

\begin{figure}[tb]
    \centering
    \includegraphics[width=0.45\textwidth]{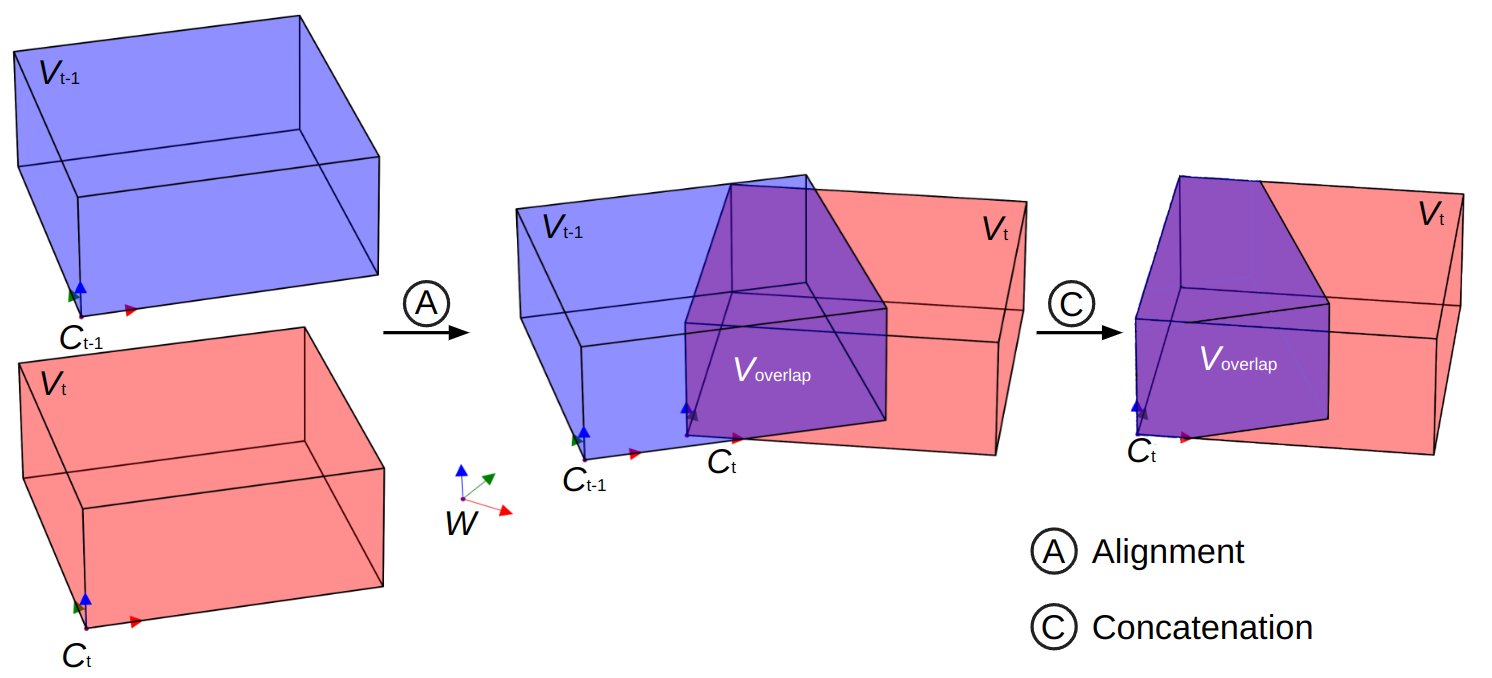}
    \caption{Concept of temporal feature propagation. In consecutive frames $C_{t-1}$ and $C_t$, the blue and red cubes represent the defined volumes, respectively. The previous volume $V_{t-1}$ is aligned to the current volume $V_{t}$ via coordinate transformation. The overlap area, denoted as $V_{overlap}$ and visualized in purple, is repeatedly estimated by neighboring frames so should be consistent. Then the feature located at the same global position is concatenated to propagate information across frames.}
    \label{fig:overlap}
    \vspace{-1.em}
\end{figure}

With known poses for each frame, we use the coordinate transformation to align features between consecutive frames. Considering the sequential camera coordinates $C_i, i = 1,\ldots,t-1,t,\dots,N$, $C_t$ refers the current frame and $C_{t-1}$ refers the previous frame, which is shown in Fig.~\ref{fig:overlap}.

Given the poses of two frames $T^{C_{t-1}}_{W}$, $T^{C_t}_{W}$ and the corresponding volume, $V_{t-1}$, $V_t$, a 3D point $\vec{P}_{C_{t-1}}$ in $V_{overlap}$ can be propagated to a corresponding point $\vec{P}_{C_t}$ in $ V_t$ via the transformation:
\begin{align}
\vec{P}_{C_t} =   T^{C_t}_{W}\cdot (T^{C_{t-1}}_{W})^{-1} \cdot \vec{P}_{C_{t-1}}.
\end{align}
Although  $\vec{P}_{C_t}$ and $\vec{P}_{C_{t-1}}$ are represented in different coordinate systems, they denote the same physical point in space. This principle of correspondence also applies to 3D features. CNNs naturally maintain the spatial structure of features, so the transformation is suitable for multiple hierarchical levels, which establishes one-to-one correspondences between features. Since this method already handles inter-frame alignment explicitly, it prevents the model from learning complex relationships caused by ego-motion. As a result, the model can leverage the 3D semantic scene representation from the previous frames as a prior to improve the current semantic scene completion.

\subsection{Losses}
\label{sec:loss}
SLCF-Net leverages multiple loss functions including the standard cross-entropy loss $L_\textrm{ce}$, the losses $L_\textrm{mono}$ proposed by MonoScene~\cite{cao2022monoscene}, and the inter-frame consistency loss $L_\textrm{con}$:
\begin{align}
L =  L_\textrm{ce} + L_\textrm{mono} + L_\textrm{con}.
\end{align}
In this study, we delve into the inter-frame consistency loss $L_\textrm{con}$ which penalizes discrepancies in the estimations within overlapping regions of adjacent frames. Specifically, we generate a pseudo ground truth using the previously estimated probability distribution and then compute the cross-entropy with the current estimated probability distribution. Furthermore, given that the ground truth is derived from real-world scans, areas obscured by occlusions remain unknown. Consequently, we only compute losses within the known regions.

\subsection{Pipeline for Training, Validation, Testing}
\label{sec:train}
SLCF-Net's pipeline for Training, Validation, and Testing is similar to RNNs. During training, due to the large memory size of the 3D data, only two consecutive frames are loaded in a single iteration. With the input sequence $\vec{x} = (x_1,\ldots,x_t,\ldots,x_T)$, estimated sequence $\hat{\vec{y}} = (\hat{y}_1,\ldots,\hat{y}_t,\ldots,\hat{y}_T)$, and hidden states $\vec{h} = (h_1,\ldots,h_t,\ldots,h_T)$, the forward process is described as follows:
\begin{align}
y_t, h_t = f(x_t, h_{t-1}).
\end{align}
For the first frame, when $t=1$, the previous hidden state $h_{t-1}$ is initialized with an initial hidden feature $h_\textrm{init}$. The model then estimates the scene as $y_1$, while simultaneously updating the hidden state $h_1$ for the subsequent frame. After processing the second frame in the same manner, gradients are propagated backward through time using Backpropagation Through Time (BPTT). Given the brevity of the sequence, the initialization of the hidden state becomes vital. Thus, we treat the initial hidden feature $h_\textrm{init}$ as a learnable parameter, enabling the model to learn the optimal initial state across the dataset during training. In our ablation study, we will also evaluate other initialization methods, such as zero initialization and random initialization. For both validation and testing, we utilize the trained $h_\textrm{init}$ to initialize $h_0$. The model then sequentially infers over the entire sequence.

\section{Evaluation}
\label{sec:Evaluation}
\begin{table*}[t]
\caption{Quantitative results on SemanticKITTI validation set.}
\vspace{-.5em}
\label{tab:evaluation}
\centering
\renewcommand{\arraystretch}{1.1} 
\linespread{0.95}\selectfont
\setlength{\tabcolsep}{1.3pt}
\sisetup{separate-uncertainty}
\begin{threeparttable}
\begin{tabular}{c|c|c|ccccccccccccccccccc|c}
  \toprule
  & &SC &\multicolumn{20}{c}{SSC}\\
   Method & Input & IoU & 
   \rotatebox{90}{car} & \rotatebox{90}{bicycle} & \rotatebox{90}{motorcycle} & \rotatebox{90}{truck} & \rotatebox{90}{other-vehicle} &
   \rotatebox{90}{person} & \rotatebox{90}{bicyclist} & \rotatebox{90}{motorcyclist} & \rotatebox{90}{road} & \rotatebox{90}{parking} &
   \rotatebox{90}{sidewalk} & \rotatebox{90}{other-ground} & \rotatebox{90}{building} & \rotatebox{90}{fence} & \rotatebox{90}{vegetation} &
   \rotatebox{90}{trunk} & \rotatebox{90}{terrain} & \rotatebox{90}{pole} & \rotatebox{90}{traffic-sign} & mIoU\\
  \midrule
   LMSCNet~\cite{roldao2020lmscnet} & x$_{occ}$ & 
   29.63 &
   18.64 & 0.00 & 0.00 & 0.00 & 0.00 & 
   0.00 & 0.00 & 0.00 & 40.69 & 4.82 &
   18.24 & 0.00 & 9.28 & 1.33 & 14.34 &
   0.02 & 18.49 & 0.00 & 0.00 & 
   6.62 \\
   
   AICNet~\cite{li2020anisotropic} & $x_{rgb}$, $x_{depth}$ &
   31.38 &
   15.34 & 0.00 & 0.00 & \underline{5.43} & 0.00 & 
   0.00 & 0.00 & 0.00 & 45.68 & 12.45 &
   21.34 & \underline{0.08} & 13.05 & 3.26 & 16.54 &
   3.21 & 29.03 & 0.07 & 0.00 &
   8.71 \\
   
   JS3C-Net~\cite{yan2021sparse} & $x_{rgb}$, $x_{pts}$ &
   \underline{39.63} &
   \underline{25.88} & 0.00 & 0.00 & 5.07 & \underline{7.38} & 
   \underline{0.74} & \underline{0.31} & 0.00 & \underline{53.01} & \textbf{14.13} &
   \underline{27.30} & 0.07 & \underline{18.04} & \underline{4.73} & \underline{19.92} &
   \underline{4.55} &  \underline{29.55} & \underline{4.34} & \underline{1.52} &
   \underline{11.34} \\
   
   \midrule
   SLCF-Net (ours) & $x_{rgb}$, $x_{depth}$ &
   \textbf{43.64} &
   \textbf{31.87} & 0.00 & \textbf{0.15} & \textbf{7.48} & \textbf{8.07} & 
   \textbf{0.84} & \textbf{0.34} & 0.00 & \textbf{57.50} & \underline{13.91} & 
   \textbf{29.65} & \textbf{0.45} & \textbf{24.39} & \textbf{12.12} & \textbf{27.66} & 
   \textbf{11.68} & \textbf{36.30} & \textbf{11.66} & \textbf{4.76} &
   \textbf{14.68} \\
   \bottomrule
\end{tabular}
\end{threeparttable}

\vspace*{1ex} 

\footnotesize{IoU for Scene Completion (SC) task and individual class mIoU for Semantic Scene Completion (SSC) task. \textbf{Best} and \underline{second best} results are highlighted.}
\vspace*{-2ex}
\end{table*}
\subsection{Experiment Setup}
\subsubsection{Dataset}
We evaluate the proposed SLCF-Net on the SemanticKITTI dataset, a real-world urban driving scenario dataset. For the SSC task, the official benchmark uses a single LiDAR scan occupancy within a predefined volume as its input. Specifically, the scene is voxelized into a $256\times256\times32$ grid with \SI{0.2}{\meter} voxels and labeled across 21 categories (19 semantic classes, 1 empty space, and 1 unknown area). The SSC dataset comprises 22 sequences: Sequences~0-7, 9-10 are for training, Sequence~8 for validation, and Sequences~11-21 for testing. The ground truth for the test set is not publicly released. Instead of that, the benchmark provides an online evaluation interface for the generated scenes on the test set but restricts the number of submissions.

Unlike the standard LiDAR occupancy input, our approach utilizes an RGB image combined with a corresponding sparse depth map. The field-of-view for this setup only covers a portion of the entire volume. To ensure a fair comparison with other SSC methods, we retrained the baseline models with the partial data in the field-of-view and also evaluated them only in the field-of-view. Due to the special assessment approach, we evaluate our approach on the original validation set, i.e. we split the dataset as follows. Sequences~0-7 are for training, Sequences~9-10 are for validation, and Sequence~8 for testing. Therefore, we did not use the hidden test set. Moreover, extensive ablation studies and visualization were conducted solely on the released validation set, Sequence~8.

\subsubsection{Training Setup}
We trained our model on an NVIDIA RTX A6000 card with 48 GB of memory. Due to the substantial memory demands of 3D data, we adopted a sequential training approach, loading only two consecutive frames of data in a single iteration. This implies that the model only utilized the information from one previous frame. The training spanned 30 epochs utilizing the AdamW optimizer with a weight decay of 1e-4. The initial learning rate was set at 1e-4, which was reduced by a factor of 10 at the 15th epoch.

\subsubsection{Metrics}
Following the benchmark, the task of scene completion is assessed using Intersection-over-Union (IoU). This metric primarily classifies a voxel as either occupied or empty. Additionally, mean IoU (mIoU) is employed for the task of semantic scene completion across 19 classes. We also apply these metrics to evaluate inter-frame consistency. Specifically, we take the estimations of the neighboring frame and calculate the IoU and mIoU in the overlap area.

\subsection{Performance}
\label{sec:performance}
Tab.~\ref{tab:evaluation} reports the performance of SLCF-Net and other SSC baselines on the SemanticKITTI validation set. For a fair comparison, the baseline method follows the official implementation but is retrained with the partial data in the field of view, which is also the area we evaluate all methods. Our method outperforms all baselines in both SC and SSC metrics. More specifically, the SLCF-Net has the highest or second-highest accuracy across all individual classes. Qualitative results for some frames are shown in Fig.~\ref{fig:visualization}.

\definecolor{carColor}{RGB}{100,150,245}
\definecolor{bicycleColor}{RGB}{100,230,245}
\definecolor{motorcycleColor}{RGB}{30,60,150}
\definecolor{truckColor}{RGB}{80,30,180}
\definecolor{othervehicleColor}{RGB}{0,0,255}
\definecolor{personColor}{RGB}{255,30,30}
\definecolor{bicyclistColor}{RGB}{255,40,200}
\definecolor{motorcyclistColor}{RGB}{150,30,90}
\definecolor{roadColor}{RGB}{255,0,255}
\definecolor{parkingColor}{RGB}{255,150,255}
\definecolor{sidewalkColor}{RGB}{75,0,75}
\definecolor{othergroundColor}{RGB}{175,0,75}
\definecolor{buildingColor}{RGB}{255,200,0}
\definecolor{fenceColor}{RGB}{255,120,50}
\definecolor{vegetationColor}{RGB}{0,175,0}
\definecolor{trunkColor}{RGB}{135,60,0}
\definecolor{terrainColor}{RGB}{150,240,80}
\definecolor{poleColor}{RGB}{255,240,150}
\definecolor{trafficsignColor}{RGB}{255,0,0}
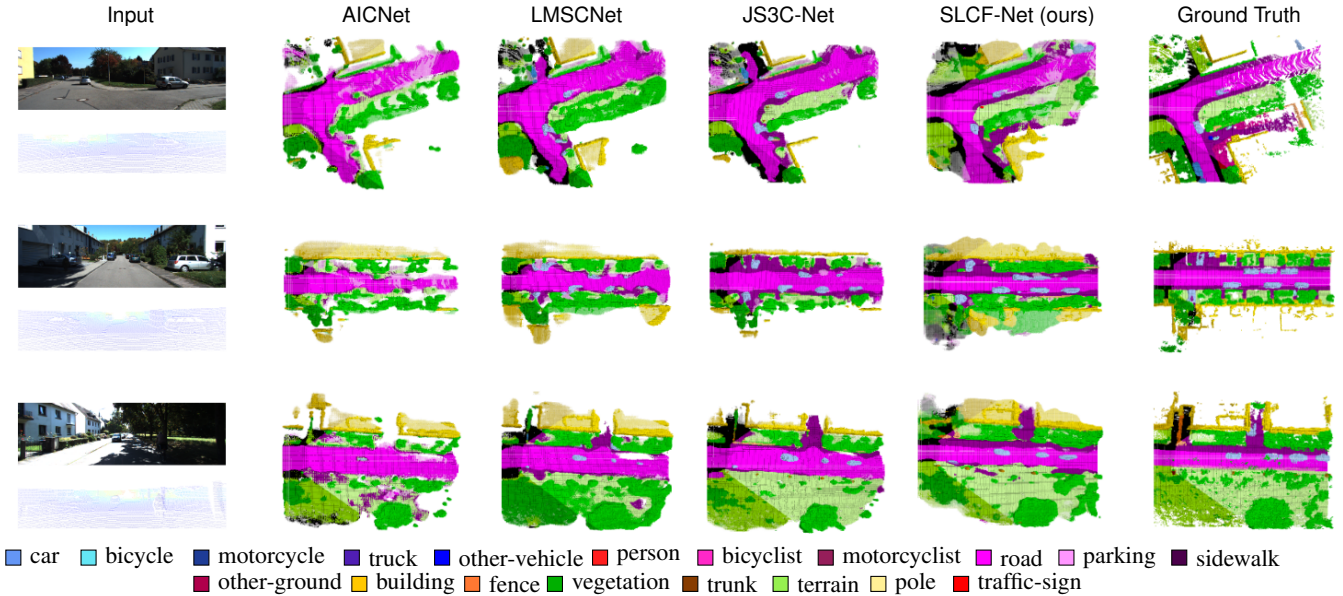
\begin{figure*}[t]
  \centering
	{\sf \footnotesize \hspace*{.8cm}Input\hspace*{2.5cm}AICNet\hspace*{1.6cm}LMSCNet\hspace*{1.6cm}JS3C-Net\hspace*{1.4cm}SLCF-Net (ours)\hspace*{1.1cm}Ground Truth}
  \begin{tikzpicture}
    \node[anchor=south west,inner sep=0] (image) at (0,0) {\includegraphics[width=\textwidth]{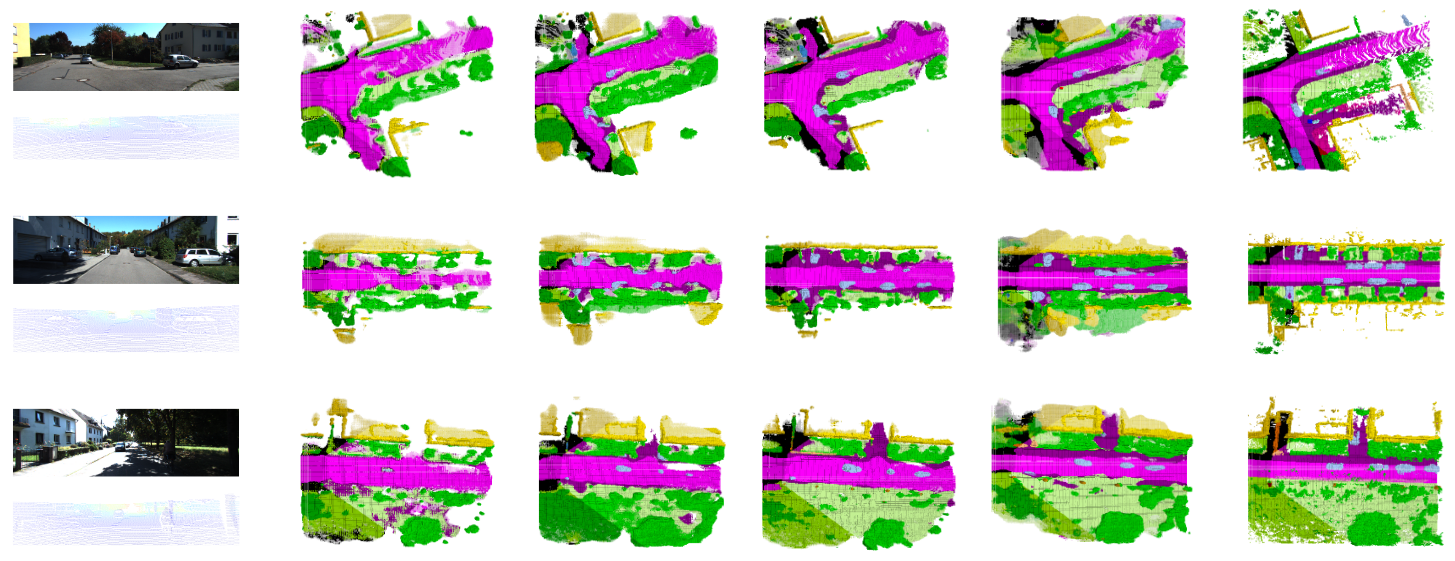}};
    \def\legendWidth{2.0}
    \def\legendHeight{0.2}
    \def\legendSpacing{0.15}
    \def\yOffset{-0.3}
    
   \draw[fill=carColor] ($ (image.south west) + (0.0, \yOffset)$) rectangle ++(0.2, 0.2) node[right, yshift=-0.1cm] {\small car};
	\draw[fill=bicycleColor] ($ (image.south west) + (1.0, \yOffset)$) rectangle ++(0.2, 0.2) node[right, yshift=-0.1cm] {\small bicycle};
    \draw[fill=motorcycleColor] ($ (image.south west) + (2.5, \yOffset)$) rectangle ++(0.2, 0.2) node[right, yshift=-0.1cm] {\small motorcycle};
    \draw[fill=truckColor] ($ (image.south west) + (4.5, \yOffset)$) rectangle ++(0.2, 0.2) node[right, yshift=-0.1cm] {\small truck};
    \draw[fill=othervehicleColor] ($ (image.south west) + (5.7, \yOffset)$) rectangle ++(0.2, 0.2) node[right, yshift=-0.1cm] {\small other-vehicle};
    \draw[fill=personColor] ($ (image.south west) + (7.8, \yOffset)$) rectangle ++(0.2, 0.2) node[right, yshift=-0.1cm] {\small person};
    \draw[fill=bicyclistColor] ($ (image.south west) + (9.2, \yOffset)$) rectangle ++(0.2, 0.2) node[right, yshift=-0.1cm] {\small bicyclist};
    \draw[fill=motorcyclistColor] ($ (image.south west) + (10.8, \yOffset)$) rectangle ++(0.2, 0.2) node[right, yshift=-0.1cm] {\small motorcyclist};
    \draw[fill=roadColor] ($ (image.south west) + (12.9, \yOffset)$) rectangle ++(0.2, 0.2) node[right, yshift=-0.1cm] {\small road};
    \draw[fill=parkingColor] ($ (image.south west) + (14.0, \yOffset)$) rectangle ++(0.2, 0.2) node[right, yshift=-0.1cm] {\small parking};
    \draw[fill=sidewalkColor] ($ (image.south west) + (15.5, \yOffset)$) rectangle ++(0.2, 0.2) node[right, yshift=-0.1cm] {\small sidewalk};
    \draw[fill=othergroundColor] ($ (image.south west) + (2.5, \yOffset-\legendHeight - \legendSpacing)$) rectangle ++(0.2, 0.2) node[right, yshift=-0.1cm] {\small other-ground};
    \draw[fill=buildingColor] ($ (image.south west) + (4.6, \yOffset-\legendHeight - \legendSpacing)$) rectangle ++(0.2, 0.2) node[right, yshift=-0.1cm] {\small building};
        \draw[fill=fenceColor] ($ (image.south west) + (6.1, \yOffset-\legendHeight - \legendSpacing)$) rectangle ++(0.2, 0.2) node[right, yshift=-0.1cm] {\small fence};
    \draw[fill=vegetationColor] ($ (image.south west) + (7.2, \yOffset-\legendHeight - \legendSpacing)$) rectangle ++(0.2, 0.2) node[right, yshift=-0.1cm] {\small vegetation};
    \draw[fill=trunkColor] ($ (image.south west) + (9.0, \yOffset-\legendHeight - \legendSpacing)$) rectangle ++(0.2, 0.2) node[right, yshift=-0.1cm] {\small trunk};
    \draw[fill=terrainColor] ($ (image.south west) + (10.2, \yOffset-\legendHeight - \legendSpacing)$) rectangle ++(0.2, 0.2) node[right, yshift=-0.1cm] {\small terrain};
    \draw[fill=poleColor] ($ (image.south west) + (11.5, \yOffset-\legendHeight - \legendSpacing)$) rectangle ++(0.2, 0.2) node[right, yshift=-0.1cm] {\small pole};
    \draw[fill=trafficsignColor] ($ (image.south west) + (12.6, \yOffset-\legendHeight - \legendSpacing)$) rectangle ++(0.2, 0.2) node[right, yshift=-0.1cm] {\small traffic-sign};
  \end{tikzpicture}
	
	\vspace*{-1ex}

  \caption{Qualitative results on SemanticKITTI. From the bird's eye view, the 19 classes are shown without empty space. The estimated voxels that are located at the unknown region are visualized with $20\%$ opacity. The region located outside of the FoV is shaded.}
  \label{fig:visualization}
  \vspace{-1.em}
\end{figure*}

\subsection{Ablation Studies}
To gain a deeper understanding of the influence of key parameters and processing steps methods in our method, we conducted a series of experiments. When investigating the impact of a specific factor, we kept other network structures and parameters consistent. We will discuss the findings in the following three aspects.

\subsubsection{Variance in GDP}
In the GDP model, the depth prior $\hat{d}$ is inferred by the network, making it not completely reliable and potentially impacting feature mapping. To address this issue, we introduced a Gaussian-decay function, enabling the model to update the depth prior during training. Among its parameters, the variance $\sigma$ is particularly influential on model performance and the learning process.

\begin{figure}[h!]
    \centering
    \begin{subfigure}[b]{0.24\textwidth}
        \includegraphics[width=\textwidth]{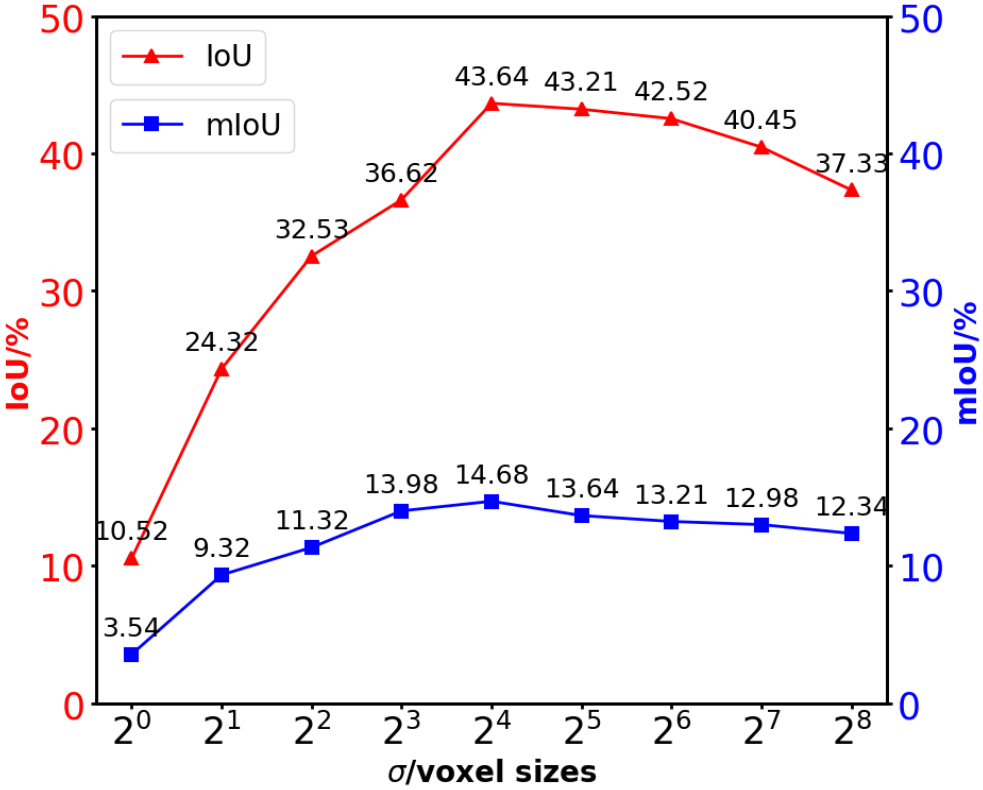}
        \caption{Model's performance}
        \label{fig:sigmaa}
    \end{subfigure}
    \hspace{0.05cm} 
    \begin{subfigure}[b]{0.22\textwidth}
        \includegraphics[width=\textwidth]{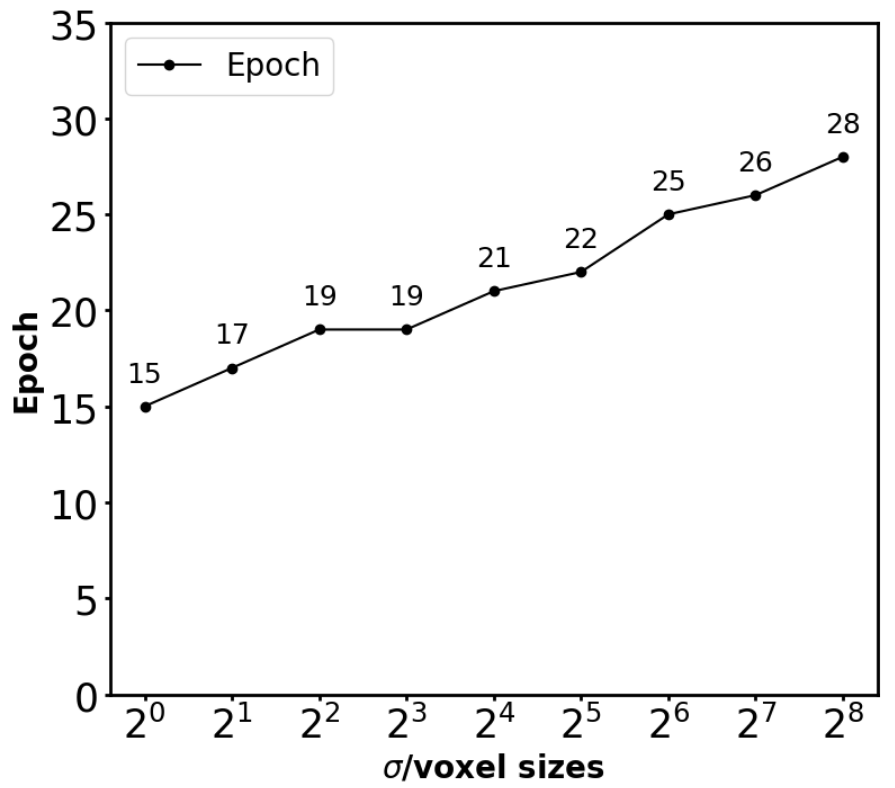}
        \caption{Learning speed}
        \label{fig:sigmab}
    \end{subfigure}
    \caption{Influence of $\sigma$ on model performance and learning process.}
    \label{fig:sigma}
     \vspace{-1.5em}
\end{figure}

Given that in SemanticKITTI the volume is defined as $256 \times 256 \times 32$, we set values for $\sigma$ as $2^n$, where $n$ ranges from 1 to 8. Specifically, when $n=0$, $ \sigma$ equals one voxel size, suggesting the model has high trust in the depth prior and is less likely to change it during training. Conversely, at $n=8$, $\sigma$ equals $256$ voxel sizes, covering the entire volume. This implies the model distributes nearly equal weights to all voxels in the line of sight, regardless of the depth prior. Fig.~\ref{fig:sigma} details the results of these experiments.

From Fig.~\ref{fig:sigmaa}, the model reaches its best performance at $\sigma=2^4$, indicating this value is the most suitable given the capability of the depth estimation module. As $\sigma$ increases beyond this, model performance starts to decline. This reveals that giving too little importance to depth priors can degrade the model's estimation ability. When $\sigma$ is less than $2^4$, the model performance significantly degrades, which signifies that errors in the depth prior can mislead the model into local optima. Further observations from Fig.~\ref{fig:sigmab} show that with the increase of $\sigma$, the epochs required for optimal performance also increase, signifying that while a larger $\sigma$ may not drastically reduce performance, it can slow down the learning process.

\subsubsection{Initial State in Hidden Layer}
In traditional RNN or LSTM architectures, both zero-initialization and random-initialization of hidden features typically wouldn't severely impact model performance, as the model can progressively learn and adjust over long sequence data. However, in the context of our work, several factors limit the length of the training sequence, making the initialization strategy profoundly impactful on model performance. We also assessed SLCF-Net's performance under zero-initialization and random-initialization policies, the results of which are reported in Tab.~\ref{tab:initial_hidden}. Compared to learnable initialization updated during training, both methods degrade performance.

\begin{table}[b!]
\vspace{-2.0em}
\caption{Evaluation of different initialization policies.}
\label{tab:initial_hidden}
\centering
\renewcommand{\arraystretch}{1.} 
\linespread{0.95}\selectfont
\setlength{\tabcolsep}{5pt}
\begin{threeparttable}
\begin{tabular}{c|cc}
  \toprule
  Method & IoU & mIoU\\
  \midrule
  zero initialization & 35.32 & 10.34\\
  random initialization & 32.64 & 9.32\\
  \midrule
  learnable initialization (ours) & \textbf{43.64} & \textbf{14.68}\\
  \bottomrule
\end{tabular}
\end{threeparttable}
\vspace{-1.5em}
\end{table}

\subsubsection{Trade-off between Accuracy and Consistency}
Using historical data can boost the model's performance, but adding a penalty for inter-frame inconsistency during training is a double-edged sword. In Tab.~\ref{tab:accuracy_consistency}, we evaluate SLCF-Net's accuracy and consistency with and without $L_\textrm{con}$, comparing it to baselines. Here, consistency is measured using IoU and mIoU in the overlap area of consecutive frames. The results show that, given the incorporation of historical data, SLCF-Net outperforms the baseline in terms of consistency. With the addition of $L_\textrm{con}$, consistency further improves, but at the cost of reduced accuracy. This trade-off arises because when there's a discrepancy between the earlier estimation and the current truth, the model has to align both. While $L_\textrm{con}$ smoothens sequence estimation, accuracies in one frame can adversely influence the subsequent frame's results.

\begin{table}[t]
\caption{Evaluation of accuracy and consistency.}
\label{tab:accuracy_consistency}
\centering
\renewcommand{\arraystretch}{1.} 
\linespread{0.95}\selectfont
\setlength{\tabcolsep}{5pt}
\begin{threeparttable}
\begin{tabular}{c|c|c|c|c}
  \toprule
  & \multicolumn{2}{c|}{Accuracy} & \multicolumn{2}{c}{Consistency}\\
  Method & IoU & mIoU & IoU & mIoU \\
  \midrule
  LMSCNet~\cite{roldao2020lmscnet}& 30.04& 6.70& 26.35& 5.34\\
  JS3C-Net~\cite{yan2021sparse}& 43.88& 11.34& 31.54& 6.75\\
  AICNet~\cite{li2020anisotropic}& 31.38& 8.71& 20.35& 4.32\\
  SLCF-Net (ours)& 39.64& 10.63& \textbf{35.62} & \textbf{8.63}\\
  SLCF-Net w/o $L_\textrm{con}$&\textbf{43.64} & \textbf{14.68}&33.65 &7.62\\
  \bottomrule
\end{tabular}
\end{threeparttable}
\vspace{-1.5em}
\end{table}

\section{Conclusions}
\label{sec:Conclusion}
For the task of semantic scene completion (SSC), we proposed SLCF-Net which leverages sequences of RGB images and sparse LiDAR depth maps as input. Central to SLCF-Net is its feature fusion mechanism, which integrates a 2D-to-3D feature projection and an inter-frame feature propagation. We tested SLCF-Net using the SemanticKITTI dataset and compared its performance with other SSC techniques.

To further improve SLCF-Net's efficacy, we conducted a series of targeted experiments. These explored the effects of the Gaussian variance, the strategy of initializing hidden features, and the implementation of an inter-frame consistency loss. Although SLCF-Net demonstrates notable advantages, it presents a trade-off between accuracy and consistency, which means effectively using historical information is still an open challenge. Additionally, the current method is designed for static environments, thereby oversimplifying the dynamic objects in real-world urban contexts. Propagating semantic scene representations using estimated scene flow~\cite{SchuttRB:ICRA22, lin2024icpflow} is a promising direction for future research.

\newpage
\IEEEtriggeratref{22}
\bibliographystyle{IEEEtran}
\bibliography{literature}

\end{document}